\documentclass{article}

\PassOptionsToPackage{breaklinks}{hyperref}

\usepackage[final, nonatbib]{neurips_2020}

\usepackage[utf8]{inputenc} 
\usepackage[T1]{fontenc}    
\usepackage{xurl}            
\usepackage{booktabs}       
\usepackage{amsfonts}       
\usepackage{nicefrac}       
\usepackage{microtype}      
\usepackage{xcolor}

\title{Diluted Near-Optimal Expert Demonstrations for Guiding Dialogue Stochastic Policy Optimisation}

\author{Thibault Cordier\textsuperscript{*,1,2}, Tanguy Urvoy\textsuperscript{2}, Lina M. Rojas-Barahona\textsuperscript{2}, Fabrice Lefèvre\textsuperscript{1} \\
  \textsuperscript{1}LIA - University of Avignon, Avignon, France \\
  \textsuperscript{2}Orange Labs, Lannion, France \\
  \texttt{thibault.cordier@alumni.univ-avignon.fr} \\
  \texttt{fabrice.lefevre@univ-avignon.fr} \\
  \texttt{\{thibault.cordier, linamaria.rojasbarahona, tanguy.urvoy\}@orange.com} \\
}

\begin{document}
\maketitle
\begin{abstract}
A learning dialogue agent can infer its behaviour from interactions with the users.
These interactions can be taken from either human-to-human or human-machine conversations. However, human interactions are scarce and costly, making learning from few interactions essential.
One solution to speedup the learning process is to guide the agent's exploration with the help of an expert. 
We present in this paper several imitation learning strategies for dialogue policy where the guiding expert is a near-optimal handcrafted policy.
We incorporate these strategies with state-of-the-art reinforcement learning methods based on Q-learning and actor-critic.
We notably propose a randomised exploration policy which allows for a seamless hybridisation of the learned policy and the expert, which can be seen as a dilution of the expert's demonstration into the resulting policy.
Our experiments show that our hybridisation strategy 
outperforms several baselines, and that it could accelerate the learning when facing real humans.
\end{abstract}


\section{Introduction} \label{sec:intro}

Spoken dialogue systems have gained more and more attention in both academy and industry.
The first spoken dialogue systems where mainly handcrafted~\cite{traum2003information}.
Nowadays, Reinforcement Learning (RL) has become a key element for designing dialogue strategies. Indeed, dialogues are modelled with Partial Observable Markov Decision Processes (POMDPs) as an optimisation problem, simulating the inherent dynamic behaviour of human conversations to find the global optimal policy~\cite{young_pomdp-based_2013}.
Furthermore, Deep RL (DRL)~\cite{li_deep_2018} has achieved significant success on many complex sequential decision-making problems and in particular in conversational AI approaches~\cite{gao_neural_2018}.
The ability to learn from few interactions is essential in dialogue applications because human interactions are scarce and costly. Unfortunately, standard RL algorithms usually require a large amount of interactions with the environment to reach good performances.

One solution to speedup the learning process is to guide the agent’s exploration.
We claim that policy learning should not be deterministic.
By taking a closer look at the way dialogue policy learning works, we observed that some actions do not contribute to the dialogue and therefore do not change the current state. 
We posit that greedy off-policy learning methods suffer this situation and need a randomised policy rather than a deterministic one. This also can be profitable for a better exploration strategy besides epsilon-greedy.
Moreover, former works did used stochastic policy (eg~\cite{young_pomdp-based_2013})
and recent methods show that stochastic learning policies like soft-kind policies can perform very well and present elegant proprieties~\cite{haarnoja_reinforcement_2017,haarnoja_soft_2018,gao_reinforcement_2019}.
The first question we address here is: can dialogue policy learning be improved with stochastic off-policy learning methods?

Secondly, we think that external demonstrations should not be reserved only for supervision and instead should be directly integrated in the reinforcement learning process.
Several methods search a way to exploit demonstrations to accelerate the learning with the conviction that demonstrations are the solution to the sparse reward. These methods are presented as useful and efficient as showed in~\cite{hester_deep_2017}. They can be based on Imitation Learning (IL) to learn an "optimal" policy function or on Inverse Reinforcement Learning (IRL) to learn an "optimal" reward function. 
Nevertheless, the majority of these approaches uses Supervised Learning (SL) and compares them with pure reinforcement methods, which is not ideal because:
(i) the integrity of policy learning is safe obviously thanks to the refinement of the policy in the reinforcement learning environment; 
(ii) a supervised update in policy learning is more powerful than a reinforced one \cite{hester_deep_2017, su_continuously_2016, gao_reinforcement_2019}. In addition, SL strongly constrains the learning of the agent according to the demonstrations and the performance of these methods therefore depends on the quality of the demonstrations.

Although warm SL can obviously improve the performance, demonstrations may still be useful to guide efficiently the exploration even when they are sub-optimal. Similarly, we think that exploration with demonstrations can improve learning.
In addition, we consider that human expertise can be used in different ways. The most classical one is to use demonstrations that humans have already produced. Another way is to use a rule-based agent, namely handcrafted, that has been designed, evaluated and fine-tuned by humans. Indeed,~\citep{casanueva_benchmarking_2017} have shown that handcrafted approaches still perform better than policy learning approaches.
Moreover it is well known that human interactions and manually crafted rules are not only costly but also time consuming; while simulated interactions are cheaper and easier to collect~\citep{su_-line_2016,schatzmann2006survey}.
Therefore, the second issue that raised in our work is: can we use demonstrations without supervision and only as a way to guide exploration in an on-line learning?

Our contribution in this paper focuses on exploration strategies in pure reinforcement learning. To the best of our knowledge, this might be the first proposal to learn a dialogue policy with demonstrations given by a sub-optimal expert directly in the learning process without any supervision.
We openly propose to learn a randomised exploration policy in order to integrate seamlessly the expert actions and to avoid being stuck on a deterministic policy. The process expert's demonstrations can be viewed as diluted in the derived policy (in contrast to bootstrap it).
Furthermore, the approaches are evaluated in the Pydial framework~\cite{ultes_pydial:_2017} on different domains and by increasing difficulty as in~\cite{casanueva_benchmarking_2017}. 
Our experimental results show that randomised exploration policies do improve performance when classical approaches fail in difficult environments and show that demonstrations given by the handcrafted expert help to improve performance at the onset of learning process without any supervision. Even more our approach tends to surpass the performance of the handcrafted expert when classical approaches fail to do that.


\section{Related Work} \label{sec:related}

On the one hand, different dialogue policy learning methods have already been tested on Pydial~\cite{casanueva_benchmarking_2017}. First of all,
GP-Sarsa~\cite{gasic_gaussian_2013} is a classical approach using Gaussian processes. 
Since then, Deep Learning approaches as DQN~\cite{mnih_playing_2013} and DDQN~\cite{van_hasselt_deep_2015} have become the new baseline for task-oriented dialogue systems and are used in the majority of recent papers. 
Then, a variety of advanced deep approaches with actor-critic have been used to improve performance such as DA2C with and without pre-supervised learning~\cite{fatemi_policy_2016},
TRACER and eNACER with and without pre-supervised learning~\cite{su_sample-efficient_2017}, 
TRACER in large action spaces context~\cite{weisz_sample_2018},
and TRACER with ICM~\cite{wesselmann_curiosity-driven_2019}.

Most of RL methods do not use a stochastic policy or sub-optimal demonstrations into the RL process directly. In contrast, RL with deep energy-based policies are promising methods for learning stochastic policies such as Soft Q-learning~\cite{haarnoja_reinforcement_2017} and Soft Actor-Critic~\cite{haarnoja_soft_2018}.
A recent method even tries to learn a stochastic policy from imperfect demonstrations but only during the pretraining phase in a video game context~\cite{gao_reinforcement_2019}.
In addition, this modelling is very closed to Boltzmann sampling exploration in discrete actions space. We are so convinced that it is the stochastic sampling which is the key of exploration-exploitation balance. In addition, using demonstrations in RL process can accelerate the exploration phase and initiate the exploitation more quickly. It justifies that these two techniques are used jointly.

On the other hand, apprenticeship learning (AL)\footnote{Apprenticeship learning (AL) (or learning from demonstration) is the process of learning by observing an expert.}
can be viewed as a form of supervised learning, where the training data set consists of task executions by a demonstration teacher~\cite{argall_survey_2009}. It is usually implemented as either IL or IRL. 
In standard IL methods, expert demonstrations can be used to guide the agent in relevant situations to be learned, such as in DAgger~\cite{ross_reduction_2011}, AggraVate~\cite{ross_reinforcement_2014} or LOLS~\cite{chang_learning_2015}.
There are different ways to integrate this paradigm into the RL process:
(i)~SL as pre-trained phase (i.e behavior cloning)~\cite{su_continuously_2016,gao_reinforcement_2019,goecks_integrating_2020};
(ii)~RL with only demonstrations as pre-training phase~\cite{hester_deep_2017,gao_reinforcement_2019,goecks_integrating_2020};
(iii)~combined SL and RL losses during RL phase~\cite{hester_deep_2017,goecks_integrating_2020, gordon-hall_learning_2020}.
More recent methods use non-standard ways to use expert demonstrations for instance
learning on single demonstration with resetting to a demo state~\cite{salimans_learning_2018}, or learning a weak and cheap expert policy for simulation~\cite{gordon-hall_learning_2020}.
In the same vein, there are different methods that use expert demonstrations in IRL: (i) engineered reward or engineered policy, for instance 
expert-based reward shaping and exploration scheme~\cite{ferreira_emmanuel_expert-based_2013}, social signal and user adaptation~\cite{ferreira_social_2013}, reward and policy shaping~\cite{wang_learning_2020} and
(ii) learned reward directly~\cite{chandramohan_user_2011,hutchison_bayesian_2014}, using deep adversarial structure~\cite{fu_learning_2018} or reward estimation/learning~\cite{takanobu_guided_2019, huang_semi-supervised_2020}.

Although these methods make interesting use of expert demonstrations, they divert the classic use of the rewards provided by the environment with supervision on significant amount of computation time. This constrains the agent to learn to do as the expert does in one way or another. But in sum, they solve the RL problem only thanks to the refinement of the policy into the RL process. As a consequence AL can be seen as a way to learn a prior policy. Conversely, we show that demonstrations can help to learn faster if they are integrated directly into the RL process without prior policy.


\section{Proposed Approach} \label{sec:approach}

The proposed approach to policy dialogue learning is based on recent deep learning methods to which we bring the advantages of stochastic techniques, namely Boltzmann sampling, and to which we integrate demonstrations directly into the RL process in order to better guide exploration and make relevant exploitation.
RL is usually implemented as either value-based method as Q-learning or policy-based method as actor-critic. Both are our baselines with which we will make our contributions.

\subsection{Preliminaries}

As a reminder of the dialogue modelling in RL, we can view the partially observable Markov decision process (POMDP) as a continuous-space Markov decision process (MDP) in terms of policy optimisation where the states are the belief states.
The belief state $b_t$ is a vector encoding a probability distribution over the different goals, dialogue acts and concepts that are discussed in the dialogue. In the same way, the dialogue action $a_t$ is a vector encoding a probability distribution over the possible agent dialogue actions.

The role of the agent is to select the dialogue action that will lead to maximize the discounted cumulative reward and so succeed the dialogue outcome with high probability. More details on dialogue modelling can be found in the benchmark described in~\cite{casanueva_benchmarking_2017} or in~\cite{young_pomdp-based_2013}. 

\subsection{Baseline Algorithms}

The Q-learning baseline is the combination of Double and Duelling Deep-Q-Network with Experience Replay (DQN/D3QN)~\cite{van_hasselt_deep_2015,wang_dueling_2016}. In short, DQN is an approximation function learning method that searches to estimate the optimal state-action value function (or Q-value). Finding an optimal Q-function is equivalent to find an optimal policy by defining a greedy policy with respect to the Q-function. 

The Double DQN architecture is an alternative DQN method to mitigate the problem of overoptimistic value estimation and the Duelling DQN architecture is for learning more efficiently by decoupling value and advantage functions. On top of that experience replay can be added which is a popular technique for reducing sample correlation and for improving sample efficiency. More technical details of each component are presented in~\ref{app:Ql}.

The actor-critic baseline is the Trust Region Policy Optimization for Actor-Critic with Experience Replay (A2C/ACER/TRACER)~\cite{wang_sample_2017,weisz_sample_2018}. In brief, an actor $\pi$ tries to maximize the expected reward when in the same time a critic $Q$ learned separately evaluates the actor decisions. This method is based on the \textit{Policy Gradient Descent}~\cite{sutton_policy_2000} for which we bring computation improvements. 

The importance sampling truncation with bias correction technique is used to correct the perceived sampling distribution induced by experience replay in order to reduce variance. We apply the Retrace algorithm~\cite{munos_safe_2016} to recursively estimate the advantage function in safe and efficient way with small bias and variance. Finally, we use the trust region policy optimization (TRPO)~\cite{schulman_trust_2017} for adjusting the policy gradient in order to learn in a safe parameters region limiting the deterioration of the policy performance. More technical details of each component are presented in~\ref{app:ACER}.

\subsection{Proposed Exploration Strategy}

We present our contributions related to the exploration strategy, which we integrate to the off-policy learning methods explained earlier. The first one proposes to learn and play a stochastic policy related to energy-based model. The second one shows how we can integrate sub-optimal expert demonstrations and feed-backs directly in the reinforcement learning process.

\subsubsection{Stochastic Exploration related to Energy-Based Function} \label{ss:explo}

Exploring with a smooth stochastic policy seems to be more adapted than exploring with a deterministic one or with $\epsilon$-greedy for instance.
We propose to learn an energy-based policy in the dialogue environment where the agent samples his actions according to Boltzmann's stochastic sampling.

\paragraph{Energy-based policies modelling:} Inspired from~\cite{haarnoja_reinforcement_2017}, we opt for using energy-based policies of the following form:

\begin{gather}
    \pi(a_t|b_t) \propto \exp(-\mathcal{E}(b_t,a_t))
\end{gather}

When using value-based method, the energy function can be represented by the Q-function with parameter $\tau$, the temperature, where we set $ \mathcal{E}(b_t,a_t) = -\frac{1}{\tau}Q^\pi(b_t,a_t)$. When using policy-based method, the energy function is directly represented by the policy network and so is learned implicitly.

In theory, learning an energy-based policy function is motivated by the consideration of the maximum entropy RL setup in which the rewards are augmented with an entropy term~\cite{haarnoja_reinforcement_2017}.
Even if there exists an interesting relationship between the temperature and the relative importance of entropy over rewards, we decide to separate the maximum entropy objective from maximum rewards objective. This results to maximizing entropy without discount factor.
In other words, we maximize the entropy at all time step without taking into account the time decay and so we promote exploration at any time step and not only at the beginning. 

\paragraph{Energy-based sampling:} A stochastic sampling with energy-based policies can be achieved by the Boltzmann sampling. Contrary to the commonly used strategy as $\epsilon$-greedy where actions $a_t$ are sampled from $(1-\epsilon) \argmax_a \pi(a|b_t) + \epsilon U(\mathcal{A})$ where $U(\mathcal{A})$ is an uniform distribution over action space, the Boltzmann sampling strategy samples actions from $\exp(-\mathcal{E}(b_t,a_t))$, hence $a_t \sim \pi(a_t|b_t)$. 

We think that exploring with smooth stochastic policy should improve the stability of learning in spite of exploitation strength. Indeed, stochastic sampling is equivalent to play current policy when greedy sampling is equivalent to play hypothetical optimal policy with respect to the current policy. 

One of its advantages is that policy learning is less influenced by the policy function changes. Conversely, the $\epsilon$-greedy sampling faces sudden jumps in action choices due to the argmax operator. So in theory, the Boltzmann strategy can make learning more stable than the $\epsilon$-greedy.

Another advantage is that the temperature parameter can control the exploration-exploitation balance. So, to counter the weakness of its exploitation, it can be interesting to define correctly the temperature. 

In practice, we decide to add random exploration in such a way that the actions are sampled according to $a_t \sim (1-\epsilon)\pi(a_t|b_t) + \epsilon U(\mathcal{A})$, namely $\epsilon$-Boltzmann sampling, with decreasing $\epsilon$ parameter in order to explore enough before following the stochastic policy with a fixed $\tau$ temperature parameter.

\subsubsection{Demonstrations Integrated Directly in the Reinforcement Learning Process}

During the reinforcement learning process, demonstrations can serve as an efficient way to explore the environment even if not optimal. Indeed, they can lead the agent to receive rewards promptly and so can lead it to exploit confident winning trajectories quickly.

We propose two demonstration sampling strategies corresponding to two different ways of using the knowledge of an adviser expert. They are both drawn on IL techniques but they are not designed in a supervised fashion. The first one is inspired by the Behaviour Cloning (BC) method and the second one by the DAgger method proposed by~\cite{ross_reduction_2011}. In this paper, we call them respectively "learning with demonstrations" and "learning with feed-backs".

\paragraph{Learning with demonstrations:} Let us assume that we learn with an offline expert i.e. demonstrations are given before learning. We propose that in $\beta$ (in percent) of dialogues, the agent plays for itself. Otherwise in $1-\beta$ of dialogues, the expert gives to the agent one of its expert trajectories as demonstration and the agent replays the dialogue as if it was the one who played it. 

This technique therefore allows us to provide the agent with promising demonstrations to quickly obtain rewards. In other point of view, the agent learns as if they are two datasets. The first one contains its trajectories and the second the expert demonstrations.

\paragraph{Learning with feed-backs:} Let us assume that we learn with an online expert i.e. demonstrations are given during learning. We propose that in $\beta$ (in percent) of dialogue actions, the agent plays for itself. Otherwise in $1-\beta$ of dialogue actions, the expert gives to the agent its expert action as feed-back and the agent plays the dialogue action as if it was its choice. 

This technique therefore allows us to provide the agent with promising dialogue actions to quickly explore the environment by playing relevant actions in a given state. In other words, it learns about its trajectories in which the expert can redirect it at any moment to more relevant action, as DAgger does.

\paragraph{Demonstrations with stochastic sampling:} In any case, when the agent plays for itself, it can play its exploration strategy. In this paper, it is the $\epsilon$-Boltzmann sampling exploration presented in~\ref{ss:explo}.

In both situations, a handcrafted agent is used to simulate near-optimal demonstrations and feed-backs. This one is not optimal as seen in the benchmark~\cite{casanueva_benchmarking_2017} but it offers good performance compared with the other deep learning methods. Also, it has been designed, evaluated and fine-tuned by humans. So it is a way to mimic human expertise and can be served as a near-optimal expert in our experimentation.


\section{Experiments} \label{sec:expres}

In our experiments the Pydial framework~\cite{ultes_pydial:_2017} is used, which implements an agenda-based user simulation~\cite{schatzmann_agenda-based_2007}. As in~\cite{casanueva_benchmarking_2017} we tested our algorithms for policy learning on different domains and in different environments by increasing the inputs' noise.
As shown in Table~\ref{tab:domains}, the domains differ from each other by the ontology size, impacting the state and action space dimensions.

\begin{table}[htbp]
\caption{Description of the domains. The third column represents the number of database search constraints that the user can define, the fourth the number of information slots the user can request from a given database entry and the fifth the sum of the number of values of each requestable slot.}
\begin{center}
\resizebox{0.75\textwidth}{!}{%
\begin{tabular}{l| c c c c}
Domain & Code & \multicolumn{1}{c}{\# constraint slots} & \multicolumn{1}{c}{\# requests}& \multicolumn{1}{c}{\# values} \\ \midrule 
Cambridge Restaurants & CR & 3 & 9 & 268\\ 
San Francisco Restaurants & SFR & 6 & 11 & 636 \\ 
Laptops & LAP & 11 & 21 & 257\\ 
\end{tabular}}
\vspace{1em}
\label{tab:domains}
\end{center}
\vspace*{-0.6cm}
\end{table}

In particular, in the scope of this paper, we decide to evaluate our policy models according to three levels of noise with respect to the semantic error rate (SER). This corresponds to the noise that comes from the ASR and the NLU channels. In Pydial, this is modelled at the semantic level whereby the true user action is corrupted by noise to generate an N-best-list with associated confidence scores. 

Table~\ref{tab:tab1} shows the compared policy models. HDC corresponds to the handcrafted policy learning, which is a rule-based approach written by experts. DQN and ACER are the baselines enhanced with stochastic (stoc) exploration and either behaviour cloning (BC) or feedbacks (FB). 

\begin{table}[hbt]
\caption{Overview of proposed methods}
\centering
\resizebox{0.65\textwidth}{!}{%
\begin{tabular}{llr}
Method name & Abbrev. \\
\midrule \midrule
Handcrafted Policy & {HDC} \\
\midrule
Stochastic Q-learning & {Stoc-DQN} \\
Stochastic Q-learning with Demonstrations & {Stoc-DQN-BC} \\
Stochastic Q-learning with Feed-backs & {Stoc-DQN-FB} \\
\midrule
Stochastic Actor-Critic & {Stoc-ACER} \\
Stochastic Actor-Critic with Demonstrations & {Stoc-ACER-BC} \\ 
Stochastic Actor-Critic with Feed-backs & {Stoc-ACER-FB} \\
\bottomrule
\end{tabular}
}
\label{tab:tab1}
\end{table}
\paragraph{Hyperparameters and Training}
All algorithms use by default $\epsilon$-Boltzmann sampling strategy during training and testing. We use common parameters between the different algorithms as learning rate ($\alpha=0.0001$), experience replay buffer size ($10000$), batch size ($128$), discount factor ($\gamma=0.9$), linearly decreasing epsilon ($\epsilon\in[0.05;0.55]$), fixed temperature ($\tau=100$), weights dropout ($0.1$), expert-agent play ratio ($\beta=0.5$). Others parameters related to referred papers are chosen identically.
All deep networks have the same architecture composed by two hidden layers with respectively $128$ and $64$ neurons. The Adam optimiser was used to train all the deep-RL models. The first hidden layer is shared between learning modules depending on the algorithms. Each network is trained after one dialogue with one training iteration.
As in~\cite{casanueva_benchmarking_2017} the maximum dialogue length was set to $25$ turns and the discount factor $\gamma$ was $0.9$. We evaluated our policy models on the the average success rate and average reward. 

\paragraph{Experiment $1$:}  
First, we performed a short training stage over $1\,000$ dialogues during which the agent learns from its interactions with the environment and potentially from demonstrations or feed-backs given by the simulated expert. 
We compare the performances with the Hard-DQN which uses a deterministic sampling and with eNAC (episodic Natural Actor-Critic) which uses the true natural policy gradient. Both are given by~\cite{casanueva_benchmarking_2017}. At this stage we search to answer the question: can we improve dialogue policy learning with demonstrations directly in the reinforcement learning process? 

\paragraph{Experiment $2$:} 
Second we performed a long training stage over $10\,000$ dialogues. This will evaluate the contribution of stochastic sampling strategy during training and testing stages.  Here we search to answer the question: can we improve dialogue policy learning with stochastic off-policy learning methods in order to compete the handcrafted agent?

All methods are evaluated after training over $1\,000$ dialogues during which the learned policy is fixed. For the second experiments, we decide to compute the average performance over the last five checkpoints from dialogue indices $6\,000$ to $10\,000$ with a step of $1\,000$ dialogues. This calculation is done in order to reduce variance induced by RL when we estimate the performance of the models.

\section{Results} 

The results of the first experiments are presented in Table~\ref{tab:results_exp}. We can observe that demonstrations can improve the performance during the early training when compared with baselines (i.e eNAC and Hard-DQN). Particularly, our proposed stochastic ACER models outperform eNAC baseline on all environments. Although stochastic ACER is already efficient to learn in early training without demonstrations, we can notice that the best performances are obtained through an expert-guided policy.

\begin{table*}[ht]
\caption{Results of Experiment $1$. Short term learning after $1\,000$ training dialogues for $1\,000$ testing dialogue. Each bold result represent the best model according to the referenced baseline.}

\centering
\resizebox{!}{0.85\height}{%
\begin{tabular}{ll||rr|rrrrrr||rr}
 &  & \multicolumn{2}{c}{eNAC} & \multicolumn{2}{c}{Stoc-ACER} & \multicolumn{2}{c}{Stoc-ACER-BC} & \multicolumn{2}{c}{Stoc-ACER-FB} & \multicolumn{2}{c}{HDC} \\ 
\multicolumn{2}{c}{\textit{Task}}  & \multicolumn{1}{c}{Suc.} & \multicolumn{1}{c}{Rew.} & \multicolumn{1}{c}{Suc.} & \multicolumn{1}{c}{Rew.} & \multicolumn{1}{c}{Suc.} & \multicolumn{1}{c}{Rew.} & \multicolumn{1}{c}{Suc.} & \multicolumn{1}{c}{Rew.} & \multicolumn{1}{c}{Suc.} & \multicolumn{1}{c}{Rew.} \\
\midrule \midrule
\multirow{3}{*}{\rotatebox[origin=c]{0}{$0\%$ SER}} 
 & CR  & 93.0\% & 12.20 & 98.5\% & \textbf{13.47} & \textbf{99.1\%} & 13.40 & 99.0\% & 13.46 & 100.0\% & 14.00 \\
 & SFR & 85.8\% &  9.90 & 97.4\% & \textbf{11.77} & \textbf{97.9\%} & 11.71 & 96.3\% & 11.50 & 98.2\%  & 12.40 \\
 & LAP & 84.2\% &  8.80 & 96.5\% & 11.65 & \textbf{98.7\%} & 12.00 & 98.4\% & \textbf{12.11} & 97.0\%  & 11.70 \\
\midrule
\multirow{3}{*}{\rotatebox[origin=c]{0}{$15\%$ SER}}
 & CR  & 85.7\% & 10.00 & 93.9\% & 11.17 & \textbf{95.9\%} & \textbf{11.72} & 94.8\% & 11.38 & 96.7\% & 11.00 \\
 & SFR & 73.6\% &  6.20 & 90.7\% & 8.88  & \textbf{92.8\%} & \textbf{9.11}  & 90.5\% & 8.84  & 90.9\% & 9.00  \\
 & LAP & 71.0\% &  5.50 & \textbf{93.2\%} & \textbf{9.60} & 83.8\% & 7.51  & 89.7\% & 9.01  & 89.6\% & 8.70  \\ 
\midrule
\multirow{3}{*}{\rotatebox[origin=c]{0}{$30\%$ SER}}
 & CR  & 73.6\% &  6.70 & 86.6\% & 8.82 & 81.6\% & 8.15 & \textbf{89.4\%} & \textbf{9.26} & 89.6\% & 9.30 \\
 & SFR & 55.2\% &  1.40 & 79.6\% & 4.62 & 81.3\% & 4.92 & \textbf{83.1\%} & \textbf{5.46} & 79.0\% & 6.00 \\
 & LAP & 56.3\% &  1.90 & \textbf{81.6\%} & \textbf{5.79} & 78.5\% & 4.67 & 79.1\% & 5.29 & 76.1\% & 5.30 \\ 
\midrule
\end{tabular}
}

\resizebox{!}{0.85\height}{%
\begin{tabular}{ll||rr|rrrrrr||rr}
 &  & \multicolumn{2}{c}{Hard-DQN} & \multicolumn{2}{c}{Stoc-DQN}  & \multicolumn{2}{c}{Stoc-DQN-BC} & \multicolumn{2}{c}{Stoc-DQN-FB} & \multicolumn{2}{c}{HDC} \\ 
\multicolumn{2}{c}{\textit{Task}}  & \multicolumn{1}{c}{Suc.} & \multicolumn{1}{c}{Rew.} & \multicolumn{1}{c}{Suc.} & \multicolumn{1}{c}{Rew.} & \multicolumn{1}{c}{Suc.} & \multicolumn{1}{c}{Rew.} & \multicolumn{1}{c}{Suc.} & \multicolumn{1}{c}{Rew.} & \multicolumn{1}{c}{Suc.} & \multicolumn{1}{c}{Rew.} \\
\midrule \midrule
\multirow{3}{*}{\rotatebox[origin=c]{0}{$0\%$ SER}} 
 & CR  & 88.6\% & 11.60 & 72.7\% & 8.08 & 93.4\% & 12.30  & \textbf{98.1\%} & \textbf{13.23} & 100.0\% & 14.00 \\
 & SFR & 48.0\% &  2.70 & 64.5\% & 3.51 & \textbf{97.2\%} & \textbf{11.56} & 93.5\% & 10.69 & 98.2\%  & 12.40 \\
 & LAP & 61.9\% & 5.50 & 57.6\% & 2.54 & \textbf{95.3\%} & \textbf{11.38} & 81.7\% & 8.41  & 97.0\%  & 11.70 \\
\midrule
\multirow{3}{*}{\rotatebox[origin=c]{0}{$15\%$ SER}}
 & CR  & 79.5\% & 9.20 & 68.4\% & 5.47  & 88.8\% & 10.17 & \textbf{92.2\%} & \textbf{10.59} & 96.7\% & 11.00 \\
 & SFR & 42.4\% & 1.00 & 56.1\% & -0.79 & \textbf{93.5\%} & \textbf{9.34}  & 87.8\% & 7.41  & 90.9\% & 9.00  \\
 & LAP & 51.9\% & 3.10 &43.7\% & -5.82 & 87.6\% & 8.06  & \textbf{88.0\%}   & \textbf{8.31} & 89.6\% & 8.70  \\ 
\midrule
\multirow{3}{*}{\rotatebox[origin=c]{0}{$30\%$ SER}}
 & CR  & 72.3\% & 6.90 & 64.9\% & 2.50  & 82.8\% & \textbf{8.17} & \textbf{83.0\%}   & 7.67 & 89.6\% & 9.30 \\
 & SFR & 35.6\% & -1.20 & 62.3\% & 1.10  & 73.5\% & 3.01 & \textbf{78.8\%} & \textbf{4.18} & 79.0\% & 6.00 \\
 & LAP & 47.5\% & 1.40 & 71.0\%   & 1.78 & 76.1\% & 4.20  & \textbf{77.1\%} & \textbf{4.45} & 76.1\% & 5.30 \\ 
\midrule
\end{tabular}
}

\label{tab:results_exp}
\end{table*}

Moreover, the proposed Stoc-DQN policy models outperform also Hard-DQN for all environments. We can also notice that demonstrations clearly improve performance of the models. Indeed, we see that demonstrations efficiently guide the exploration of the models. For instance, Stoc-DQN-BC model outperforms all the others for SFR and laptops in the environment ($0\%$ SER) and for SFR in environment ($15\%$ SER). Similarly, Stoc-DQN-FB outperforms all the other models for all domains in the environment ($30\%$ SER).
This attests a faster exploitation of the environment that therefore improves outcomes in more difficult environments.

Finally, we can observe that in some cases the stochastic policy models outperform the handcrafted expert, such as: Stoc-ACER with $15\%$ SER and $30\%$ SER for laptops; Stoc-ACER-BC with $0\%$ SER for laptops and with $15\%$ SER for SFR; Stoc-ACER-FB with $30\%$ SER applied SFR. It is worth noting that laptops contains more constrains than the other domains (Table~\ref{tab:domains}). In addition, with $0\%$ and $15\%$ SER the the performance of Stoc-ACER-BC is very close to the expert for CR. This suggest that integrating demonstrations directly in the RL process without supervision can improve dialogue policy learning.

\begin{table*}[ht]
\caption{Results of Experiment $2$. Long term learning, average from $6\,000$ to $10\,000$ training dialogues, for $1\,000$ testing dialogue. Each bold result represent better models than the handcrafted agent.}
\centering
\resizebox{!}{0.8\height}{%
\begin{tabular}{ll||rrrrrr||rr}
 &  & \multicolumn{2}{c}{Stoc-DQN}  & \multicolumn{2}{c}{Stoc-DQN-BC} & \multicolumn{2}{c}{Stoc-DQN-FB}  & \multicolumn{2}{c}{HDC} \\ 
\multicolumn{2}{c}{\textit{Task}}  & \multicolumn{1}{c}{Suc.} & \multicolumn{1}{c}{Rew.} & \multicolumn{1}{c}{Suc.} & \multicolumn{1}{c}{Rew.} & \multicolumn{1}{c}{Suc.} & \multicolumn{1}{c}{Rew.} & \multicolumn{1}{c}{Suc.} & \multicolumn{1}{c}{Rew.} \\
\midrule \midrule
\multirow{3}{*}{\rotatebox[origin=c]{0}{$0\%$ SER}} 
 & CR  & 97.48\% & 12.67 & 98.34\% & 13.30 & 98.88\% & 13.46 & \textbf{100.0\%} & \textbf{14.00} \\
 & SFR & 94.18\% & 10.99 & 87.40\% & 9.58  & 95.62\% & 10.94 & \textbf{98.2\%}  & \textbf{12.40} \\
 & LAP & 95.08\% & 11.10 & \textbf{98.10\%} & \textbf{11.76} & \textbf{98.40\%} & \textbf{11.77} & 97.0\%  & 11.70 \\
\midrule
\multirow{3}{*}{\rotatebox[origin=c]{0}{$15\%$ SER}}
 & CR  & 92.22\% & 10.29 & 94.16\% & 11.26 & 95.76\% & \textbf{11.64} & \textbf{96.7\%} & 11.00 \\
 & SFR & 90.64\% & 8.56  & 88.70\% & 8.06  & 89.22\% & 8.21  & \textbf{90.9\%} & \textbf{9.00}  \\
 & LAP & \textbf{90.34\%} & 8.59  & \textbf{92.56\%} & \textbf{9.40}  & \textbf{91.86\%} & \textbf{9.19} & 89.6\% & 8.70 \\
\midrule
\multirow{3}{*}{\rotatebox[origin=c]{0}{$30\%$ SER}}
 & CR  & 84.32\% & 7.73 & 85.36\% & 8.64 & 85.46\% & 8.65 & \textbf{89.6\%} & \textbf{9.30} \\
 & SFR & \textbf{82.48\%} & 5.11 & \textbf{80.26\%} & 5.05 & \textbf{80.34\%} & 4.63 & 79.0\% & \textbf{6.00} \\
 & LAP & \textbf{82.24\%} & \textbf{5.89} & \textbf{84.62\%} & \textbf{6.25} & \textbf{83.40\%} & \textbf{6.00}  & 76.1\% & 5.30 \\ 
\midrule
\end{tabular}
}
\resizebox{!}{0.8\height}{%
\begin{tabular}{ll||rrrrrr||rr}
 &  & \multicolumn{2}{c}{Stoc-ACER} & \multicolumn{2}{c}{Stoc-ACER-BC} & \multicolumn{2}{c}{Stoc-ACER-FB} & \multicolumn{2}{c}{HDC} \\ 
\multicolumn{2}{c}{\textit{Task}}  & \multicolumn{1}{c}{Suc.} & \multicolumn{1}{c}{Rew.} & \multicolumn{1}{c}{Suc.} & \multicolumn{1}{c}{Rew.} & \multicolumn{1}{c}{Suc.} & \multicolumn{1}{c}{Rew.} & \multicolumn{1}{c}{Suc.} & \multicolumn{1}{c}{Rew.} \\
\midrule \midrule
\multirow{3}{*}{\rotatebox[origin=c]{0}{$0\%$ SER}} 
 & CR  & 99.60\% & \textbf{14.02} & 99.64\% & \textbf{14.03} & 99.30\% & 13.87 & \textbf{100.0\%} & 14.00 \\
 & SFR & 97.66\% & 12.36 & 96.48\% & 11.98 & 96.34\% & 11.98 & \textbf{98.2\%}  & \textbf{12.40} \\
 & LAP & 95.24\% & 11.23 & 95.34\% & 11.28 & 95.40\% & 11.24 & \textbf{97.0\%}  & \textbf{11.70} \\
\midrule
\multirow{3}{*}{\rotatebox[origin=c]{0}{$15\%$ SER}}
 & CR  & \textbf{97.56\%} & \textbf{12.69} & 95.80\%  & \textbf{12.32} & \textbf{96.78\%} & \textbf{12.59} & 96.7\% & 11.00 \\
 & SFR & 88.62\% & \textbf{9.26}  & 87.64\% & 8.91  & 86.98\% & 8.67  & \textbf{90.9\%} & 9.00  \\
 & LAP & 88.74\% & \textbf{8.72}  & 88.00\% & 8.55  & 86.26\% & 8.24  & \textbf{89.6\%} & 8.70 \\
\midrule
\multirow{3}{*}{\rotatebox[origin=c]{0}{$30\%$ SER}}
 & CR  & \textbf{89.82\%} & \textbf{10.19} & 89.38\% & \textbf{9.98} & 89.32\% & \textbf{10.04} & 89.6\% & 9.30 \\
 & SFR & 72.74\% & 4.38  & 78.22\% & 5.45 & 71.02\% & 4.37  & \textbf{79.0\%} & \textbf{6.00} \\
 & LAP & 75.80\% & 4.86  & \textbf{78.66\%} & \textbf{5.40} & \textbf{77.82\%} & 5.23  & 76.1\% & 5.30 \\ 
\midrule
\end{tabular}
}
\label{tab:results_stoch}
\end{table*}

The results of our second experiments are presented in Table~\ref{tab:results_stoch}. In most environments, methods learn very well compared to the benchmarks~\citet{casanueva_benchmarking_2017}. Furthermore, some of them can compete with the handcrafted agent whether it is a stochastic Q-learning approach or a stochastic actor-critic approach. For instance, Stoc-DQN-BC outperforms handcrafted expert with $30\%$ SER for laptops and SFR. Again stochastic ACER was more robust to the different environments, showing better performance, particularly for Stoc-ACER-BC with $30\%$ SER for laptops.

These results are encouraging and suggest that stochastic sampling makes it possible to learn a policy that performs as well as the handcrafted agent, which has been designed and fine-tuned by humans, even in hard environments. Also, these results show that demonstrations can significantly contribute to improve the performance at early learning stages.

\section{Conclusion and Future Work} \label{sec:conclusion}

We have presented in this paper hybridisation strategies in pure reinforcement learning, in which we learn a stochastic dialogue policy with demonstrations given by a sub-optimal expert designed by humans directly in the learning process. 
The results of our experiments suggest that this hybridisation approach outperforms classical approaches in noisy environments for task-oriented dialogue systems. Moreover, our results suggest that the demonstrations given by the handcrafted expert improve performance at the beginning of the learning process without any supervision. Our approach can compete with the handcrafted expert when classical approaches can not.

\section{Future Work} \label{subsec:futurework}

We would like to explore strategies which can help the agent to consult efficiently the demonstrations or the feed-backs of the expert by adding constraints in a similar approach of~\cite{chen_agent-aware_2017}. We would like to experiment different strategies of how define correctly temperature in Boltzmann sampling when learning with Q-learning in order to adapt these methods to take account of the action space size. Finally, we would like to train the model with different beta parameters or with a mixture of demonstrations and feed-backs because it would be an important factor in practice to estimate the importance of the diluted expert in the derived policy and so the human cost.


\section*{Acknowledgments}
This work was partially funded by the ANR Cifre convention N°2019/2006 and by Orange-Labs Research. 
The team DATA-AI/NADIA at Orange-Labs earned unlimited access to the French HPC Jean Zay (IDRIS-CNRS)\footnote{\url{http://www.idris.fr/annonces/annonce-jean-zay.html}} with the project 10096 selected in the French contest \textbf{Grands Challenges IA 2019}. Thus all the experiments presented in this paper were run in Jean Zay HPC.

\bibliography{neurips_2020}
\bibliographystyle{acl_natbib}

\newpage
\appendix
\section{Appendices} \label{sec:appendix}

\subsection{Reinforcement Learning}

Consider an finite-horizon discounted Markov decision process (MDP), defined by the tuple $(\mathcal{S}, \mathcal{A}, P, r, \rho_0, \gamma)$, where $\mathcal{S}$ is a continuous states space, $\mathcal{A}$ is a finite actions space, $P : \mathcal{S} \times \mathcal{A} \times \mathcal{S} \to \mathbb{R}$ is the transition probability distribution, $r : \mathcal{S} \to \mathbb{R}$ is the reward function, $\rho_0 : \mathcal{S} \to \mathbb{R}$ is the distribution of the initial state $b_0$, and $\gamma \in (0, 1)$ is the discount factor.

Let $\pi$ denote a policy $\pi : \mathcal{S}\times\mathcal{A}\to[0,1]$ and let $\eta(\pi)$ denote its expected discounted reward:

\begin{gather}
\eta(\pi) = \mathbb{E}\left[\sum_{t=0}^T \gamma^{t}R_t\right] \label{eq:eta} \\
\mathrm{where}\quad b_0 \sim \rho_0(B_0),\, a_t\sim\pi(A_t|B_t),\,b_{t+1}\sim P(B_{t+1}|B_t,A_t),\, r_t\sim R(B_{t+1})\nonumber
\end{gather}

The goal of the agent is to learn an optimal policy $\pi^*$ such as:

\begin{equation}
    \pi^* = \argmax_\pi \; \eta(\pi)
    \label{eq:pimax}
\end{equation}

\subsection{Baseline Algorithms}

\subsubsection{DQN as Q-learning Baseline} \label{app:Ql}

The Q-learning baseline is the Double and Dueling Deep-Q-Network with Experience Replay (DQN/D3QN)~\cite{van_hasselt_deep_2015,wang_dueling_2016} 
DQN is an approximation function learning method that searches to estimate the optimal state-action value function (or Q-value) defined as follows:

\begin{gather}
Q^*(b_t, a_t) = \argmax_\pi Q^\pi(b_t, a_t) \label{eq:qmax} \\
\mathrm{where} \quad Q^\pi(b_t, a_t) = \mathbb{E}_\pi\left[\sum_{t' \geq t} \gamma^{t'-t}R_{t'}|B_t=b_t,\,A_t=a_t\right]\nonumber
\end{gather}

Find an optimal Q-function in Eq. \ref{eq:qmax} is equivalent to find an optimal policy in Eq. \ref{eq:pimax} by defining a greedy policy with respect to the Q-function.
To estimate this function, we optimize an online network $Q(b_t,a_t;\omega_i)$ at iteration $i$ by minimizing the following loss function:

\begin{gather}
L(\omega_i) = \mathbb{E}_{b_t,a_t\sim\pi}\left[\left(q_i-Q(b_t, a_t; \omega_{i})\right)^2\right] \\
\mathrm{with} \quad q_i = r_t + \gamma \max_{a'} Q(b_{t+1}, a'; \omega_{-})\nonumber
\end{gather}

where $q_i$ is the one-step bootstrapped estimator of the Q-function and $Q(b_t,a_t;\omega_{-})$ is a target network with freeze and separate parameters $\omega_{-}$ for a fixed number of iterations in order to improve the stability of the algorithm.

The Double DQN architecture is an alternative DQN method to mitigate the problem of overoptimistic value estimation. Indeed, the operations of action selection and evaluation are done by the same Q-function. So to separate both operations, the bootstrapped estimator $q_i$ is redefined as:

\begin{gather}
q_i = r_t + \gamma\, Q(b_{t+1}, a'; \omega_{-}) \\
\mathrm{where} \quad a' = \argmax_{a'}\; Q(b_{t+1}, a'; \omega_i)\nonumber
\end{gather}

The Dueling DQN architecture is also an alternative DQN method to learn more efficiently by decoupling value and advantage functions in DQN architecture while sharing a common feature learning module. The value and advantage functions are defined as follows:

\begin{gather}
    V^\pi(b_t) = \mathbb{E}_\pi\left[\sum_{t' \geq t} \gamma^{t'-t}R_{t'}|B_t=b_t\right] \quad \mathrm{or} \quad  V^\pi(b_t) = \sum_a \pi(a|b_t)Q^\pi(b_t,a) \label{eq:Vvalue} \\
    A^\pi(b_t, a_t) = Q^\pi(b_t, a_t) - V^\pi(b_t)
    \label{eq:Avalue}
\end{gather}

In contrast of the standard DQN when only one Q-value of one action is updated and the others are untouched, every update of any Q-values in the dueling architecture results in an update of the state value at the same time. This allows a better approximation of the state values.

Each network has a common feature learning module $\omega^{(F)}$ which is divided into two modules $\omega^{(V)}$ and $\omega^{(A)}$ for value and advantage respectively. The Q-value estimation is carried out as follows:

\begin{equation}
Q(b_t, a_t; \omega) = V(b_t; \omega^{(F)},\omega^{(V)}) + \\ \left( A(b_t, a_t; \omega^{(F)},\omega^{(A)}) - \frac{1}{|\mathcal{A}|} \sum_{a'} A(b_t, a'; \omega^{(F)},\omega^{(A)}) \right)
\end{equation}

The advantage function estimator is forced to have zero advantage on average in order to address the issue of identifiability of the Q-value function in Eq. \ref{eq:Avalue} and so recover unique $V$ and $A$ functions.

Experience Replay is a popular technique for reducing sample correlation and for improving sample efficiency. In short, the system stores the agent's experiences. During learning, the Q-values updates are done on samples of experiences drown uniformly at random from the pool of stored samples. Finally, this value gradient has the following form: 

\begin{gather}
\nabla_{\omega_i}L(\omega_i) = \mathbb{E}_{b_t \sim d^\mu, a_t \sim \mu}\left[\frac{1}{2}\nabla_{\omega_i}Q(b_t, a_t; \omega_{i})\left(q_i-Q(b_t, a_t; \omega_{i})\right)\right]\\
\mathrm{with} \quad q_i = r_t + \gamma Q(b_t, a'; \omega')\nonumber\\
\mathrm{where} \quad a' = \argmax_{a'}\; Q(b_t, a'; \omega_i)\nonumber
\end{gather}

\subsubsection{ACER as Actor-Critic Baseline} \label{app:ACER}

The actor-critic baseline is the Trust Region Policy Optimization for Actor-Critic with Experience Replay (A2C/ACER/TRACER)~\cite{wang_sample_2017,weisz_sample_2018}.
In brief, an actor $\pi$ tries to maximize the expected reward Eq. \ref{eq:eta} when in the same time a critic $Q$ learned separately evaluate the actor decisions.
To estimate them, we optimize two online networks $\pi(a_t|b_t;\omega^{(F)},\omega^{(P)})$ and $Q(b_t,a_t;\omega^{(F)},\omega^{(Q)})$. They have a common feature learning module $\omega^{(F)}$ which is divided into two modules $\omega^{(P)}$ and $\omega^{(Q)}$ for policy and Q-values respectively.

According to the \textit{Policy Gradient Descent} \cite{sutton_policy_2000}, the gradient of the parameters given the objective function $J(\omega) = \eta(\pi_\omega)$ has the following form:

\begin{gather}
    \nabla_{\omega} J(\omega) = \mathbb{E}_{b_t,a_t \sim\pi}\left[\nabla_{\omega} \log \pi(a_t|b_t;\omega^{(F)},\omega^{(P)})\, Q(b_t,a_t;\omega^{(F)},\omega^{(Q)}) \right]
\end{gather}

The importance sampling technique can be used to correct the perceived sampling distribution induced by experience replay. Indeed, the behaviour policy $\mu$ (used to generate the data during learning) and the learned policy $\pi$ (used to be improved and be evaluate during final testing) are different. The importance sampling weights $\rho(a_t,b_t)=\frac{\pi(a_t,b_t)}{\mu(a_t,b_t)}$ rectifies the unbalanced sampling of interactions gathered according to $\mu$ with respect to $\pi$. However, because these weights are potentially unbounded and can result in significant variance, a truncation with bias correction technique can also be used as a countermeasure.

In order to reduce variance, the advantage function can be used in place of the Q-function as an unbiased estimate as shown in Eq. \ref{eq:Vvalue} and \ref{eq:Avalue}. Moreover, in off-policy context, we can use the Retrace algorithm \cite{munos_safe_2016} to estimate the advantage function recursively in safe and efficient way with small bias and variance by introducing eligibility traces $c_t = \lambda\min(1,\rho(a_t|b_t))$ as following:

\begin{gather}
    A^\pi_{ret}(b_t,a_t) = Q^\pi_{ret}(b_t,a_t) - V^{\pi}(b_t) \label{eq:Aret_ACER}\\[1ex]
    \mathrm{where} \quad Q^\pi_{ret}(b_t,a_t) = r_t + \gamma V^{\pi}(b_{t+1}) + \gamma c_{t+1} (Q^\pi_{ret}(b_{t+1},a_{t+1}) - Q^{\pi}(b_{t+1},a_{t+1})) \nonumber
\end{gather}

In sum, by introducing all previous techniques, the policy gradient has the following form:

\begin{gather}
    \begin{split}
        \nabla_\omega J(\omega) = \mathbb{E}_{b_t \sim d^\mu}\Bigg[
            \mathbb{E}_{a_t \sim \mu}\left[ 
                \bar\rho(a_t|b_t)\nabla_\omega \log \pi(a_t|b_t;\omega^{(F)},\omega^{(P)})\, A_{ ret}(b_t,a_t;\omega^{(F)},\omega^{(Q)}) 
            \right]\\
            + \mathbb{E}_{a \sim \pi}\left[ 
                \hat\rho(a|b_t) \nabla_\omega \log \pi(a|b_t;\omega^{(F)},\omega^{(P)})\, A(b_t,a;\omega^{(F)},\omega^{(Q)}) 
            \right]
        \Bigg]
    \end{split}\\
    \mathrm{with} \quad \bar\rho(a_t|b_t) = \min(c,\rho(a_t|b_t)) \quad \mathrm{and} \quad \hat\rho(a|b_t) = \left[\frac{\rho(a|b_t)-c}{\rho(a|b_t)}\right]_{+}\nonumber
\end{gather}

The value gradient then has the following form:

\begin{gather}
    \nabla_\omega L(\omega) = \mathbb{E}_{b_t \sim d^\mu, a_t \sim \mu}\Bigg[ \frac{1}{2}
            \nabla_\omega Q(b_t,a_t;\omega^{(F)},\omega^{(Q)}) \left(q_i - Q(b_t,a_t;\omega^{(F)},\omega^{(Q)})\right)
    \Bigg]\\
    \mathrm{with} \quad q_i = Q_{ret}(b_t,a_t;\omega^{(F)},\omega^{(Q)}) \nonumber
\end{gather}

To discourage the algorithm from learning a trivial policy, we subtract 1\% of the policy entropy from the final loss function.

Finally, trust region policy optimization (TRPO) method \cite{schulman_trust_2017} adjusts the policy gradient in order to learn in a safe parameters region. This avoids to produce small changes in the parameters space that can lead to erratic changes in the output policy and thus lead to unstable learning. It approximates the natural gradient method by restricting the Kullback-Leibler divergence between the current policy $\pi_\omega$ parametrised by $\omega$ and the updated policy parametrised by $\omega+\alpha\nabla_\omega J(\omega)$ for learning rate $\alpha$. In other words, it restricts the current policy $\pi_\omega$ to stay around a more stable average policy $\pi_{\omega_{-}}$. This average policy is updated softly after each learning step as $\omega_{-} \leftarrow \beta\omega_{-}+(1-\beta)\omega$. After considering the following optimization problem in Eq. \ref{eq:pbTRPO} by \cite{schulman_trust_2017}, the approximation gradient policy has a closed-form solution presented in Eq. \ref{eq:solTRPO}:

\begin{gather}
    \underset{z}{\mathrm{minimize}} \quad \frac{1}{2}\|g-z\|^2_2 \label{eq:pbTRPO}\\
    \mathrm{subject\; to} \quad k^Tz\leq \delta \nonumber\\
    \mathrm{with} \quad g = \nabla_\omega J(\omega) \quad \mathrm{and} \quad k = \nabla_\omega \mathrm{KL}[\pi(.|\omega_{-})\|\pi(.|\omega)] \quad \mathrm{and} \quad \delta = \mathrm{cst} \nonumber
\end{gather}

\begin{gather}
    z = g - \max\left(0,\frac{k^Tg-\delta}{\|k\|^2_2}k\right)
    \label{eq:solTRPO}
\end{gather}

\end{document}